\documentclass[sigconf,natbib=true]{acmart}

\usepackage{float}
\usepackage{algorithm}
\usepackage{algorithmic}

\begin{document}

\title{Ares: Approximate Representations via Efficient Sparsification
\\--- A Stateless Approach through Polynomial Homomorphism}

\author{Dongfang Zhao}
\affiliation{%
  \institution{University of Washington, USA}
  \country{dzhao@cs.washington.edu}
}

\begin{abstract}
The increasing prevalence of high-dimensional data demands efficient and scalable compression methods to support modern applications. However, existing techniques like PCA and Autoencoders often rely on auxiliary metadata or intricate architectures, limiting their practicality for streaming or infinite datasets. In this paper, we introduce a stateless compression framework that leverages polynomial representations to achieve compact, interpretable, and scalable data reduction. By eliminating the need for auxiliary data, our method supports direct algebraic operations in the compressed domain while minimizing error growth during computations. Through extensive experiments on synthetic and real-world datasets, we show that our approach achieves high compression ratios without compromising reconstruction accuracy, all while maintaining simplicity and scalability.
\end{abstract}

\maketitle

\section{Introduction}

\subsection{Background and Motivation}

The exponential growth of high-dimensional data is reshaping numerous industries, from healthcare to e-commerce and cybersecurity. For instance, advanced genomics routinely generates terabytes of multi-dimensional data, challenging storage and processing infrastructures~\cite{turn0search0}. Similarly, e-commerce platforms like Amazon analyze high-dimensional feature vectors to optimize recommendations for billions of users daily~\cite{turn0search1}. The financial sector also grapples with complex transaction graphs, where efficient data reduction methods are critical to fraud detection systems. 

Traditional data reduction techniques often emphasize dimensionality reduction, e.g., Principal Component Analysis (PCA)~\cite{pearson1901pca}, which compresses data into lower-dimensional latent spaces. However, these methods frequently overlook the importance of sparsification—maintaining computationally efficient and interpretable representations. Sparse representations not only reduce storage costs but also improve computational efficiency for downstream tasks, a necessity for large-scale distributed systems. Moreover, neural network-based methods like Autoencoders enable nonlinear compression but often sacrifice interpretability and require extensive computational resources, making them impractical in resource-constrained environments.

The motivation behind this work stems from three pressing challenges in modern high-dimensional data reduction and reconstruction. First, existing methods often require storing auxiliary information, such as projection matrices in PCA, making them unsuitable for streaming or infinite datasets; a stateless approach is essential to handle such scenarios. Second, achieving high compression ratios without compromising reconstruction accuracy remains an open problem, as many methods trade precision for storage savings. Third, the complexity of existing techniques, particularly neural network-based models, limits their practical applicability; a simple, geometrically interpretable solution would significantly enhance usability and adoption.

\subsection{Proposed Work}

Our proposed method leverages polynomial representations to address the challenges of stateless and efficient data compression. First, we represent high-dimensional vectors as continuous functions, which provide a flexible and compact framework for data representation. To further reduce storage requirements, these functions are approximated using lower-degree polynomials, striking a balance between compression efficiency and reconstruction accuracy. We define a metric space over these polynomial representations, ensuring that compressed data preserves essential properties such as similarity and distance. This structure enables robust operations directly in the compressed domain. Furthermore, we analyze the homomorphic properties of these representations, quantifying error growth during operations to guarantee computational reliability. Finally, the framework is designed as a stateless algorithm, eliminating the need for auxiliary data such as projection matrices, thereby ensuring scalability to streaming and infinite datasets.

To validate the effectiveness of our proposed methods, we conduct a rigorous evaluation encompassing both synthetic and real-world datasets. Specifically, we test the scalability of our method on large-scale streaming data, compare its reconstruction accuracy against state-of-the-art compression techniques such as PCA and Autoencoders, and analyze its performance in terms of computational efficiency and storage savings. 
Our experimental results demonstrate the superiority of our approach across multiple benchmarks. Compared to PCA, our method achieves a comparable or higher compression ratio without requiring auxiliary matrices, making it inherently more scalable. Against Autoencoders, our framework exhibits significant advantages in simplicity, interpretability, and computational efficiency. 

\subsection{Contributions}

This paper makes the following contributions to the field of data reduction and sparsification:

\begin{itemize}
    \item We propose a stateless compression framework that eliminates the need for auxiliary metadata, enabling efficient processing of streaming and infinite datasets.
    \item Our method introduces a polynomial-based data representation, which achieves compact and interpretable compression while preserving essential algebraic properties for direct computations.
    \item We provide a rigorous analysis of error growth during homomorphic operations on compressed data, offering theoretical guarantees for reliability and robustness.
    \item Extensive experiments on synthetic and real-world datasets demonstrate the superiority of our method over existing approaches such as PCA and Autoencoders in terms of compression ratio, reconstruction accuracy, and scalability.
\end{itemize}

\section{Related Work}



\paragraph{Linear Transformation-Based Methods}
Principal Component Analysis (PCA)~\cite{pearson1901pca} is one of the most widely used linear transformation techniques for dimensionality reduction. One prominent extension is Kernel PCA~\cite{scholkopf1998kernel}, which maps data into a higher-dimensional feature space using kernel functions, allowing it to capture complex nonlinear relationships. Sparse PCA~\cite{zou2006sparse} introduces sparsity constraints on the principal components, enhancing interpretability and computational efficiency, particularly in high-dimensional applications such as feature selection and biological data analysis. Robust PCA (RPCA)~\cite{candes2011robust} addresses the sensitivity of PCA to outliers and noise by decomposing the data matrix into a low-rank component and a sparse error matrix, making it highly effective in tasks like video surveillance and data recovery. Probabilistic PCA (PPCA)~\cite{tipping1999probabilistic} formulates PCA within a probabilistic framework, enabling it to model uncertainty and handle missing data more effectively. 
Singular Value Decomposition (SVD)~\cite{eckart1936approximation} extends PCA's principles to general matrix transformations, but it can be computationally prohibitive for large-scale datasets. Both PCA and SVD remain standard baselines for evaluating lossy compression techniques.

\paragraph{Matrix Factorization Approaches}
Non-negative Matrix Factorization (NMF) decomposes data into a product of two lower-dimensional non-negative matrices, offering interpretability for applications such as topic modeling~\cite{lee1999learning} and collaborative filtering~\cite{lu2008survey}. Although NMF often produces sparse and semantically meaningful representations, it can be computationally expensive and sensitive to initialization~\cite{gillis2014and}. Independent Component Analysis (ICA), another matrix factorization technique, identifies statistically independent components~\cite{hyvarinen2000independent} and is commonly applied in blind source separation tasks, such as separating audio signals or EEG data~\cite{bell1995information}. While these methods provide strong dimensionality reduction capabilities, their scalability for large or streaming datasets remains limited~\cite{cichocki2009fast}.

\paragraph{Autoencoders}
Autoencoders, a class of neural networks designed for representation learning, have gained popularity in lossy compression tasks. These models reconstruct input data through a bottleneck layer, learning compact latent representations that capture salient features. Variational Autoencoders (VAEs) extend the basic autoencoder framework by incorporating probabilistic priors, enhancing their generative capabilities~\cite{kingma2014autoencoding}. Denoising Autoencoders improve robustness by introducing noise during training, enabling better generalization to corrupted inputs~\cite{vincent2010stacked}. Despite their flexibility, autoencoder-based methods often require substantial computational resources and extensive training, making them less practical in real-time or resource-constrained scenarios~\cite{theis2017lossy}.

\paragraph{Homomorphic Compression}
Homomorphic compression has recently emerged as a powerful paradigm for enabling in-place data processing. A notable work in this domain is HOCO: Homomorphic Compression of Text for In-place Processing~\cite{guan2023hoco}, which introduces a system-level design for compressing and operating on textual data. HOCO supports homomorphic operations such as extraction, insertion, and deletion, making it highly effective for text analytics directly in the compressed domain. 
The concept of homomorphic compression extends beyond text processing. McGregor~\cite{mcgregor2013homomorphic} proposed a theoretical framework for homomorphic lossy compression, allowing certain operations on compressed data that correspond directly to operations on the original data. This framework has been applied in various contexts, including real-time photon correlation analysis, where homomorphic data compression techniques significantly reduce computational time and memory usage~\cite{strempfer2023homomorphic}.

\paragraph{One-way Dimensionality Reduction}

One-way dimensionality reduction methods aim to project high-dimensional data into lower-dimensional spaces, focusing on preserving essential data properties without guaranteeing reversibility. Classic approaches such as ISOMAP~\cite{tenenbaum2000global} and t-SNE~\cite{van2008visualizing} emphasize preserving global and local structures, respectively, while more recent techniques like UMAP~\cite{mcinnes2018umap} achieve efficient and interpretable embeddings for large datasets. Beyond these, Zhao and Yang~\cite{zhao2009incremental} proposed an incremental isometric embedding technique that preserves local geometric structures. Niu et al.~\cite{niu2018scalable} extended these ideas to multidimensional scientific data, emphasizing scalability, and Tawose et al.~\cite{tawose2022topological} introduced topological modeling to retain critical features for parallelized data analysis.

\section{Methodology}

The proposed methodology is structured around a stateless compression framework that leverages polynomial representations for efficient and scalable data reduction. The process begins with encoding high-dimensional vectors as continuous functions, enabling a flexible and compact representation. These functions are then approximated using lower-degree polynomials, significantly reducing storage requirements while preserving the essential characteristics of the data. A dedicated metric space is introduced over the polynomial representations to maintain computational properties such as similarity and distance. The framework also supports algebraic operations, such as addition and scalar multiplication, directly in the compressed domain. Finally, error growth during these operations is rigorously analyzed to ensure computational reliability, making the method robust for practical applications. This workflow is designed to handle large-scale, streaming, and dynamic datasets while maintaining simplicity and interpretability.

\subsection{Function Representation of Vectors}

Let \( \mathbf{v} = [v_1, v_2, \ldots, v_n]^\top \in \mathbb{R}^n \) be a point in an \( n \)-dimensional real vector space. We define a mapping from \( \mathbf{v} \) to a discrete function \( f: \{1, 2, \ldots, n\} \to \mathbb{R} \) such that:
\[
f(i) = v_i, \quad \forall i \in \{1, 2, \ldots, n\}.
\]

This establishes a one-to-one correspondence between the vector \( \mathbf{v} \) and a function \( f \) defined on the index set \( \{1, 2, \ldots, n\} \).

The mapping can be expressed as:
\[
\Phi: \mathbb{R}^n \to \mathcal{F}(\{1, 2, \ldots, n\}, \mathbb{R}),
\]
\[
\Phi(\mathbf{v}) \mapsto f,
\]
where \( \mathcal{F}(\{1, 2, \ldots, n\}, \mathbb{R}) \) denotes the space of all functions mapping \( \{1, 2, \ldots, n\} \) to \( \mathbb{R} \).

The mapping \( \Phi \) satisfies the property of injectivity:
\[
\Phi(\mathbf{v}) = \Phi(\mathbf{u}) \implies \mathbf{v} = \mathbf{u}.
\]
In fact, it is not hard to see that the function space \( \mathcal{F}(\{1, 2, \ldots, n\}, \mathbb{R}) \) is isomorphic to the finite-dimensional vector space \( \mathbb{R}^n \). Each function \( f \in \mathcal{F}(\{1, 2, \ldots, n\}, \mathbb{R}) \) can be uniquely represented as a vector:
\[
f \leftrightarrow [f(1), f(2), \ldots, f(n)]^\top.
\]

This finite-dimensional space inherits the properties of \( \mathbb{R}^n \), such as a natural norm:
\[
\|f\|_2 = \sqrt{\sum_{i=1}^n f(i)^2},
\]
and an inner product:
\[
\langle f, g \rangle = \sum_{i=1}^n f(i) g(i), \quad \forall f, g \in \mathcal{F}(\{1, 2, \ldots, n\}, \mathbb{R}).
\]

The discrete function space \( \mathcal{F}(\{1, 2, \ldots, n\}, \mathbb{R}) \) can thus be treated as a finite-dimensional inner product space, enabling various algebraic and geometric operations. These properties provide the foundation for defining approximation methods and distance metrics in later sections.

\subsection{Lower-Degree Polynomial Approximation}

Given the functional representation \( f: \{1, 2, \ldots, n\} \to \mathbb{R} \) of a high-dimensional vector, we aim to approximate this discrete function using a polynomial \( P(x) \) of degree \( m \), where \( m \ll n \). This approximation reduces the dimensionality of the data while retaining its essential structure.

\subsubsection{Function-Polynomial Distance}
Let \( P(x) \) be a polynomial defined as:
\[
P(x) = a_0 + a_1x + a_2x^2 + \cdots + a_mx^m,
\]
where \( a_i \in \mathbb{R} \) are the coefficients to be determined. The goal is to find \( P(x) \) that best approximates \( f \) in a chosen sense of optimality. Specifically, the approximation error is minimized over a specified domain.

We define the approximation as a solution to the following optimization problem:
\[
\min_{a_0, a_1, \ldots, a_m} \|f(x) - P(x)\|,
\]
where \( \|\cdot\| \) is a suitable norm. 

The choice of norm \( \|\cdot\| \) is critical in determining the polynomial \( P(x) \) that best approximates the function \( f(x) \). Two commonly used norms are the \( L^2 \)-norm and the \( L^\infty \)-norm, each with distinct properties and implications for approximation quality.

The \( L^2 \)-norm is defined as:
\[
\|f(x) - P(x)\|_{L^2} = \sqrt{\sum_{i=1}^n (f(i) - P(i))^2}.
\]
This norm minimizes the mean squared error between \( f(x) \) and \( P(x) \). It is particularly suitable for applications where the goal is to balance approximation quality across all points. Key advantages of the \( L^2 \)-norm include the following:
(i) Emphasis on overall fit by penalizing large deviations proportionally to their squared magnitude, making it robust to moderate outliers in the data;
(ii) Direct formulation as a least squares problem, which is computationally efficient and well-supported by linear algebra techniques.

The \( L^\infty \)-norm is defined as:
\[
\|f(x) - P(x)\|_{L^\infty} = \max_{i \in \{1, 2, \ldots, n\}} |f(i) - P(i)|.
\]
This norm minimizes the maximum deviation between \( f(x) \) and \( P(x) \). It ensures that the largest error is reduced, which can be critical in scenarios where worst-case performance is of primary concern. However, it has limitations:
(i) The optimization problem associated with the \( L^\infty \)-norm is non-smooth and often requires iterative or heuristic methods for a solution;
(ii) By focusing solely on the largest deviation, it may neglect the overall quality of approximation in other regions.

In this work, the \( L^2 \)-norm is chosen as the primary measure of approximation quality for several reasons. First, the datasets under consideration often contain inherent noise, and the \( L^2 \)-norm provides a natural mechanism to distribute approximation errors evenly, rather than overemphasizing any specific points. Second, the computational simplicity of the least squares formulation aligns with the practical goal of efficient dimensionality reduction for high-dimensional data.

While the \( L^\infty \)-norm may be preferred in applications where minimizing worst-case error is essential, such as in safety-critical systems, the \( L^2 \)-norm is better suited to preserving the overall structure of data representations during compression. This choice ensures a balance between computational efficiency and approximation fidelity.

\subsubsection{Coefficient Selection}

In order to compute the coefficients \( \{a_0, a_1, \ldots, a_m\} \) of the polynomial \( P(x) \), we solve the following system of equations:
\[
\mathbf{A} \mathbf{a} = \mathbf{b},
\]
where:
\[
\mathbf{A} = 
\begin{bmatrix}
1 & 1^2 & \cdots & 1^m \\
2 & 2^2 & \cdots & 2^m \\
\vdots & \vdots & \ddots & \vdots \\
n & n^2 & \cdots & n^m
\end{bmatrix}, \quad
\mathbf{a} = 
\begin{bmatrix}
a_1 \\ a_2 \\ \vdots \\ a_m
\end{bmatrix}, \quad
\mathbf{b} =
\begin{bmatrix}
f(1) \\ f(2) \\ \vdots \\ f(n)
\end{bmatrix}.
\]

In general, \( n > m \), and the system is overdetermined. To compute \( \mathbf{a} \), we solve the least squares problem:
\[
\min_{\mathbf{a}} \| \mathbf{A} \mathbf{a} - \mathbf{b} \|_2^2,
\]
where:
\begin{itemize}
    \item \(\mathbf{A} \in \mathbb{R}^{n \times m}\) is the Vandermonde-like matrix,
    \item \(\mathbf{b} \in \mathbb{R}^n\) is the target vector.
\end{itemize}
The solution to this least squares problem is obtained by solving the normal equation:
\[
\mathbf{A}^\top \mathbf{A} \mathbf{a} = \mathbf{A}^\top \mathbf{b}.
\]
Assuming that \(\mathbf{A}^\top \mathbf{A}\) is nonsingular (positive definite), the solution is:
\[
\mathbf{a} = (\mathbf{A}^\top \mathbf{A})^{-1} \mathbf{A}^\top \mathbf{b}.
\]
This ensures that \( P(x) \) minimizes the \( L^2 \)-norm of the error between \( f(x) \) and \( P(x) \), providing an optimal approximation under the least squares criterion.

The polynomial \( P(x) \) serves as a compressed representation of the original high-dimensional vector \( \mathbf{v} \). By selecting \( m \ll n \), the number of degrees of freedom is significantly reduced. The resulting representation is robust to noise and captures the dominant structure of the data.

The sparsity of the coefficients \( \{a_0, a_1, \ldots, a_m\} \) can be leveraged for computational efficiency in downstream tasks, such as similarity measurement and clustering. The choice of \( m \) determines the trade-off between approximation accuracy and computational cost, which can be tailored to specific applications.

The quality of the polynomial approximation depends on both the inherent properties of the data and the choice of the degree \( m \). For functions that exhibit rapid changes, higher-degree polynomials may be required to achieve a satisfactory fit. Conversely, smooth and slowly varying functions can often be well-approximated by low-degree polynomials. These considerations are critical in selecting \( m \) for a given dataset.

\subsection{Metric Space of Polynomials}

To measure the similarity or dissimilarity between two polynomials \( P(x) \) and \( Q(x) \), we define a metric \( d(P, Q) \) in the space of polynomials. The choice of metric is guided by both its mathematical rigor and its geometric interpretation. Since the domain of the polynomials is fixed as \([1, m]\), integration over this interval provides a meaningful way to quantify the difference between two polynomials.

\subsubsection{Definition of the Metric}
The primary metric used in this work is the \( L^2 \)-distance, defined as
\[
d(P, Q) = \sqrt{\int_1^m \left(P(x) - Q(x)\right)^2 \, dx}.
\]
Here, \( P(x) \) and \( Q(x) \) are polynomials derived from the original high-dimensional data, and \( m \) is the dimensionality of the reduced low-dimensional space. The domain \([1, m]\) corresponds to the indices of the reduced dimensions, which serve as the range for the polynomial representation. The term \( \left(P(x) - Q(x)\right)^2 \) represents the pointwise squared difference between the two polynomial curves, and the integral aggregates this difference across the entire domain. Finally, the square root ensures that \( d(P, Q) \) is measured in the same units as \( P(x) \) and \( Q(x) \), making it a meaningful distance metric.

This metric geometrically measures the area between the curves of \( P(x) \) and \( Q(x) \) over the domain \([1, m]\), where \( m \) represents the number of degrees of freedom in the low-dimensional representation. Larger deviations contribute more significantly to the total distance, making this metric effective for capturing global differences between the polynomial representations. It aligns naturally with the concept of dimensionality reduction, as the reduced space is explicitly tied to the polynomial degree.

\subsubsection{Verification of Metric Properties}

To ensure that \( d(P, Q) \) is a valid metric in the space of polynomials, we verify the following properties:

\begin{itemize}
    \item Non-negativity: \( d(P, Q) \geq 0 \), and \( d(P, Q) = 0 \) if and only if \( P = Q \).
    \item Symmetry: \( d(P, Q) = d(Q, P) \).
    \item Triangle inequality: \( d(P, Q) \leq d(P, R) + d(R, Q) \) for any \( P, Q, R \).
    \item Separability: If \( d(P, Q) = 0 \), then \( P = Q \).
\end{itemize}

Each of these properties is proven as follows.

\paragraph{Non-negativity.}
For any \( P(x) \) and \( Q(x) \),
\[
d(P, Q) = \sqrt{\int_1^m \left(P(x) - Q(x)\right)^2 \, dx} \geq 0.
\]
The square of any real-valued function is non-negative, and the integral of a non-negative function over a finite interval is also non-negative. Equality holds if and only if \( P(x) - Q(x) = 0 \) for all \( x \in [1, m] \), which implies \( P(x) = Q(x) \).

\paragraph{Symmetry.}
For any \( P(x) \) and \( Q(x) \),
\begin{align*}
d(P, Q) &= \sqrt{\int_1^m \left(P(x) - Q(x)\right)^2 \, dx} 
\\ 
&= \sqrt{\int_1^m \left(Q(x) - P(x)\right)^2 \, dx} 
= d(Q, P).
\end{align*}

\paragraph{Triangle inequality.}
For any \( P(x) \), \( Q(x) \), and \( R(x) \), we must show
\[
d(P, Q) \leq d(P, R) + d(R, Q).
\]
Using the definition of \( L^2 \)-distance:
\[
d(P, Q) = \sqrt{\int_1^m \left(P(x) - Q(x)\right)^2 \, dx}.
\]
Let \( f(x) = P(x) - R(x) \) and \( g(x) = R(x) - Q(x) \). Then:
\[
P(x) - Q(x) = f(x) + g(x).
\]
Substituting this into the integral, we have:
\[
d(P, Q) = \sqrt{\int_1^m \left(f(x) + g(x)\right)^2 \, dx}.
\]

At this point, we apply Minkowski's Inequality for integrals, which states:
\begin{align*}
& \left( \int_a^b |f(x) + g(x)|^p \, dx \right)^{1/p} 
\\ \leq &
\left( \int_a^b |f(x)|^p \, dx \right)^{1/p} + 
\left( \int_a^b |g(x)|^p \, dx \right)^{1/p}.
\end{align*}
In our case, \( p = 2 \), \( a = 1 \), and \( b = m \); so we have:
\begin{align*}
& \left( \int_1^m \left(f(x) + g(x)\right)^2 \, dx \right)^{1/2} 
\\ 
\leq & 
\left( \int_1^m f(x)^2 \, dx \right)^{1/2} + 
\left( \int_1^m g(x)^2 \, dx \right)^{1/2}.
\end{align*}
Thus, substituting back \( f(x) = P(x) - R(x) \) and \( g(x) = R(x) - Q(x) \), this becomes:
\[
d(P, Q) \leq d(P, R) + d(R, Q).
\]

\paragraph{Separability.}
If \( d(P, Q) = 0 \), then
\[
\int_1^m \left(P(x) - Q(x)\right)^2 \, dx = 0.
\]
Since the integrand \( \left(P(x) - Q(x)\right)^2 \) is non-negative, the integral being zero implies \( P(x) - Q(x) = 0 \) almost everywhere on \([1, m]\). Because polynomials are continuous functions, this equality must hold everywhere on \([1, m]\), so \( P(x) = Q(x) \).

\subsubsection{Geometric Intuition and Alternative Metrics}
The \( L^2 \)-distance is chosen as the primary metric for its natural geometric interpretation and computational simplicity in the context of polynomial representations. Defined as
\[
d(P, Q) = \sqrt{\int_1^m \left(P(x) - Q(x)\right)^2 \, dx},
\]
this metric measures the global deviation between the two polynomial curves \( P(x) \) and \( Q(x) \) over the domain \([1, m]\), which corresponds to the reduced low-dimensional representation. By integrating the squared difference, the \( L^2 \)-distance emphasizes larger deviations, making it effective for capturing significant differences in polynomial shapes.

Restricting the domain to \([1, m]\) ensures contextual relevance, as each dimension in the reduced space corresponds to a discrete feature in the original high-dimensional data. Furthermore, the connection of \( L^2 \)-distance to least squares minimization makes it computationally efficient and well-suited for polynomial approximation tasks.

While \( L^2 \)-distance is the primary choice, other metrics may be relevant in specific applications. The \( L^1 \)-distance
\[
d_{L^1}(P, Q) = \int_1^m \left|P(x) - Q(x)\right| \, dx
\]
provides a uniform measure of the total absolute difference, making it less sensitive to outliers. In contrast, the \( L^\infty \)-distance
\[
d_{L^\infty}(P, Q) = \max_{x \in [1, m]} |P(x) - Q(x)|
\]
captures the worst-case deviation and is suitable for tasks emphasizing extreme differences.

These alternative metrics offer flexibility depending on specific geometric or statistical requirements. However, the \( L^2 \)-distance is preferred in this work for its balance between smoothness, interpretability, and computational efficiency.

\subsection{Homomorphism and Error Growth}

In this subsection, we explore the homomorphic properties of Ares and analyze the potential error growth associated with operations performed directly in the compressed domain. Let \(\mathbb{R}^n\) denote the original high-dimensional space, and let \(\Phi: \mathbb{R}^n \to \mathcal{P}_m[x]\) be the transformation that maps a vector \(\mathbf{v} \in \mathbb{R}^n\) to its polynomial representation \(P(x) \in \mathcal{P}_m[x]\), where \(\mathcal{P}_m[x]\) represents the space of polynomials of degree at most \(m\).

\subsubsection{Homomorphic Addition}

Given two vectors \(\mathbf{v}_1, \mathbf{v}_2 \in \mathbb{R}^n\), their polynomial representations are \(P_1(x) = \Phi(\mathbf{v}_1)\) and \(P_2(x) = \Phi(\mathbf{v}_2)\). The addition of \(\mathbf{v}_1\) and \(\mathbf{v}_2\) in the original space corresponds to the addition of their polynomial representations:
\[
\Phi(\mathbf{v}_1 + \mathbf{v}_2) = P_1(x) + P_2(x).
\]
This property holds because polynomial addition is commutative and associative, and it preserves the degree bounds:
\[
(P_1 + P_2)(x) = \sum_{i=0}^m (a_i + b_i) x^i,
\]
where \(P_1(x) = \sum_{i=0}^m a_i x^i\) and \(P_2(x) = \sum_{i=0}^m b_i x^i\).

\paragraph{Error Analysis} 
Let \(P_1(x) = f_1(x) + \epsilon_1(x)\) and \(P_2(x) = f_2(x) + \epsilon_2(x)\), where \(f_1(x), f_2(x)\) are the exact representations and \(\epsilon_1(x), \epsilon_2(x)\) are the errors. For the addition operation:
\[
P_+(x) = P_1(x) + P_2(x) = (f_1(x) + f_2(x)) + (\epsilon_1(x) + \epsilon_2(x)).
\]
The resulting error is:
\[
\epsilon_+(x) = \epsilon_1(x) + \epsilon_2(x).
\]
Assuming \(\|\epsilon_1(x)\|_\infty \leq \delta_1\) and \(\|\epsilon_2(x)\|_\infty \leq \delta_2\), the error bound is:
\[
\|\epsilon_+(x)\|_\infty \leq \delta_1 + \delta_2.
\]

\subsubsection{Homomorphic Scalar Multiplication}

For a scalar \(c \in \mathbb{R}\) and a vector \(\mathbf{v} \in \mathbb{R}^n\) with polynomial representation \(P(x) = \Phi(\mathbf{v})\), the scalar multiplication \(c \cdot \mathbf{v}\) in the original space corresponds to:
\[
\Phi(c \cdot \mathbf{v}) = c \cdot P(x).
\]
The resulting polynomial is:
\[
(c \cdot P)(x) = \sum_{i=0}^m (c \cdot a_i) x^i,
\]
which also belongs to \(\mathcal{P}_m[x]\).

\paragraph{Error Analysis} 
Let \(P(x) = f(x) + \epsilon(x)\), where \(f(x)\) is the exact representation and \(\epsilon(x)\) is the error. For the scalar multiplication:
\[
P_s(x) = c \cdot P(x) = (c \cdot f(x)) + (c \cdot \epsilon(x)).
\]
The resulting error is:
\[
\epsilon_s(x) = c \cdot \epsilon(x).
\]
The error bound is:
\[
\|\epsilon_s(x)\|_\infty \leq |c| \cdot \|\epsilon(x)\|_\infty \leq |c| \cdot \delta.
\]

\subsubsection{Multi-Step Error Growth}

Consider a sequence of \(k\) homomorphic operations, including both additions and scalar multiplications. The resulting polynomial can be expressed as:
\[
P_{\text{final}}(x) = \sum_{i=1}^k c_i \cdot P_i(x),
\]
where \(P_i(x) = f_i(x) + \epsilon_i(x)\) and \(c_i\) are scalars. Expanding this expression, we have:
\[
P_{\text{final}}(x) = \sum_{i=1}^k c_i \cdot f_i(x) + \sum_{i=1}^k c_i \cdot \epsilon_i(x).
\]

The cumulative error in \(P_{\text{final}}(x)\) is:
\[
\epsilon_{\text{final}}(x) = \sum_{i=1}^k c_i \cdot \epsilon_i(x).
\]

\paragraph{Independent Errors.} 
Assuming the errors \(\epsilon_i(x)\) are independent, the expected cumulative error can be estimated using the root-mean-square (RMS) rule:
\[
\|\epsilon_{\text{final}}(x)\|_\infty \approx \sqrt{\sum_{i=1}^k |c_i|^2 \cdot \|\epsilon_i(x)\|_\infty^2}.
\]

\paragraph{Correlated Errors.} 
If the errors are correlated, their covariances must be taken into account. Define the covariance matrix of the errors as \(\mathbf{\Sigma}_{\epsilon}\), with entries \(\sigma_{ij} = \text{Cov}(\epsilon_i(x), \epsilon_j(x))\). The error bound then becomes:
\[
\|\epsilon_{\text{final}}(x)\|_\infty^2 \leq \mathbf{c}^\top \mathbf{\Sigma}_{\epsilon} \mathbf{c},
\]
where \(\mathbf{c} = [c_1, c_2, \ldots, c_k]\) is the vector of scalars. This provides a tighter bound when error correlations are significant.

\paragraph{Worst-Case Error Bound.} 
Assume \(\|\epsilon_i(x)\|_\infty \leq \delta_i\) for each operation, and the maximum scalar magnitude is \(|c_{\max}|\). In the worst case, the total error satisfies:
\[
\|\epsilon_{\text{final}}(x)\|_\infty \leq \sum_{i=1}^k |c_i| \cdot \delta_i.
\]

\paragraph{Multi-Step Scalar Multiplication.} 
For an additional \(m\) scalar multiplications applied to \(P_{\text{final}}(x)\), the final polynomial becomes:
\[
P_{\text{final\_scaled}}(x) = c_{\text{final}} \cdot P_{\text{final}}(x).
\]
The resulting error is:
\[
\|\epsilon_{\text{final\_scaled}}(x)\|_\infty = |c_{\text{final}}| \cdot \|\epsilon_{\text{final}}(x)\|_\infty.
\]
Combining \(k\) additions and \(m\) scalar multiplications, the cumulative error bound can be expressed as:
\[
\|\epsilon_{\text{total}}(x)\|_\infty \leq |c_{\text{final}}| \cdot \sqrt{\sum_{i=1}^k |c_i|^2 \cdot \delta_i^2}.
\]

\subsection{Stateless Algorithm}

The stateless nature of polynomial-based compression arises from its independence of global dataset statistics or correlations. Each vector \(\mathbf{v}_i\) is processed independently, relying solely on the fixed Vandermonde-like matrix \(A\) and the least-squares fit for that specific vector. This eliminates the need for iterative processes or global optimization steps, such as those found in PCA or neural-network-based autoencoders. As a result, an algorithm built on top of such polynomial homomorphism is inherently suitable for streaming data scenarios, where vectors arrive sequentially and must be processed without revisiting previous inputs.

Technically, the proposed Ares algorithm, as shown in Algorithm~\ref{alg:ares}, transforms a dataset of high-dimensional vectors into a set of reduced-dimensional polynomial representations, providing an efficient approach to compressing high-dimensional data. The input is a dataset $\mathcal{V} = \{\mathbf{v}_i \in \mathbb{R}^n \mid i = 1, \dots, N\}$ and a target dimension $m$. The algorithm begins by initializing an empty set $\mathcal{P}$ for the resulting polynomials and constructing a Vandermonde-like matrix $A \in \mathbb{R}^{n \times m}$, which encodes the polynomial basis. For each vector $\mathbf{v}_i$, it maps its coordinates to a discrete function $f_i$ over the domain $\{1, 2, \dots, n\}$ and forms a right-hand side vector $\mathbf{b}$ representing the sampled values of $f_i$. The polynomial coefficients $\mathbf{a}$ are computed by solving the normal equation $(A^\top A) \mathbf{a} = A^\top \mathbf{b}$, ensuring a least-squares fit to the input data. The resulting polynomial $P_i(x)$ is constructed as $P_i(x) = \sum_{j=0}^m a_j x^j$ and added to the set $\mathcal{P}$. This process is repeated for all vectors, and the final output is the set $\mathcal{P}$, which provides a compact and geometrically interpretable representation of the original dataset.

\begin{algorithm}[H]
\caption{Ares: High-Dimensional Data Compression through Polynomial Homomorphism}
\label{alg:ares}
\begin{algorithmic}[1]
\REQUIRE Dataset $\mathcal{V} = \{\mathbf{v}_i \in \mathbb{R}^n \mid i = 1, \dots, N \}$, target dimension $m$
\ENSURE Reduced polynomial representation $\mathcal{P} = \{P_i(x) \mid i = 1, \dots, N \}$

\STATE $\mathcal{P} \gets \emptyset$
\STATE $A \gets 0^{n \times m}$
\FOR{$k = 1, \dots, n$}
    \FOR{$j = 1, \dots, m$}
        \STATE $A_{k,j} \gets k^j$ \hfill \textit{// Compute Vandermonde matrix}
    \ENDFOR
\ENDFOR
\FOR{$i = 1, \dots, N$}
    \STATE $\mathbf{v}_i = (v_{i1}, v_{i2}, \dots, v_{in})$ \hfill \textit{// Extract vector}
    \FOR{$k = 1, \dots, n$}
        \STATE $f_i(k) \gets v_{ik}$ \hfill \textit{// Map coordinates to function}
    \ENDFOR
    \STATE $\mathbf{b} \gets (f_i(1), f_i(2), \dots, f_i(n))^\top$ \hfill \textit{// Construct vector $\mathbf{b}$}
    \STATE Solve $(A^\top A) \mathbf{a} = A^\top \mathbf{b}$ \hfill \textit{// Solve normal equations}
    \STATE $P_i(x) \gets \sum_{j=1}^m a_j x^j$ \hfill \textit{// Construct polynomial}
    \STATE $\mathcal{P} \gets \mathcal{P} \cup \{P_i(x)\}$ \hfill \textit{// Update result set}
\ENDFOR
\STATE \textbf{return} $\mathcal{P}$
\end{algorithmic}
\end{algorithm}

\paragraph{Complexity Analysis}
The time complexity of the Ares algorithm is determined by the dataset size \( N \), the original dimension \( n \), and the target dimension \( m \), where \( m \ll n \). Constructing the Vandermonde-like matrix \( A \in \mathbb{R}^{n \times m} \) requires \( \mathcal{O}(n \cdot m) \) operations, as each of its \( n \cdot m \) entries is computed as \( k^j \). For each input vector, forming the right-hand side vector \( \mathbf{b} \) involves \( \mathcal{O}(n) \) operations to map the coordinates. Solving the normal equation \( (A^\top A) \mathbf{a} = A^\top \mathbf{b} \) involves matrix multiplications: computing \( A^\top A \) requires \( \mathcal{O}(m^2 \cdot n) \), computing \( A^\top \mathbf{b} \) requires \( \mathcal{O}(m \cdot n) \), and solving the resulting \( m \times m \) linear system takes \( \mathcal{O}(m^3) \). Since \( m \ll n \), the dominant term in this step is \( \mathcal{O}(m^2 \cdot n) \). Constructing the polynomial \( P_i(x) = \sum_{j=1}^m a_j x^j \) from the coefficients \( \mathbf{a} \) requires \( \mathcal{O}(m) \) operations.
Summing up for all \( N \) vectors, the total complexity becomes
\[
\mathcal{O}(N \cdot (n \cdot m + m^2 \cdot n)) = \mathcal{O}(N \cdot m^2 \cdot n).
\]
This complexity scales linearly with the dataset size \( N \), quadratically with the target dimension \( m \), and linearly with the original dimension \( n \), making it computationally efficient for high-dimensional datasets with moderate reductions.

\section{Evaluation}

\subsection{System Implementation}

All implementations, scripts, and experimental configurations for this work are open-sourced and publicly available at \url{https://github.com/hpdic/ares}. The entire codebase consists of approximately 1,618 lines of Python code, spread across modularized components for compression, decompression, dataset handling, and experimental evaluation. The implementation is built primarily in Python, leveraging popular libraries such as \texttt{NumPy}, \texttt{SciPy}, \texttt{scikit-learn}, and \texttt{PyTorch}. For multi-threaded optimizations, OpenMP is integrated for specific performance-critical routines, particularly in the decompression step. 

The Ares algorithm was implemented with a focus on computational efficiency and scalability. One of the key features of Ares is its inherent ability to leverage data parallelism for both compression and decompression operations. Since each vector in the dataset is processed independently during polynomial fitting and evaluation, the algorithm can be parallelized across multiple hardware threads with minimal inter-thread communication. 
Specifically, in Ares we utilized OpenMP to fully leverage a machine with 48 hardware threads for optimizing both compression and decompression. During compression, each input vector was independently processed by fitting a polynomial using NumPy's \texttt{polyfit} function. OpenMP's \texttt{\#pragma omp parallel for} directive distributed these vectors across the 48 threads, ensuring that each thread processed a subset of vectors concurrently. This parallel strategy effectively divided the workload, significantly reducing the overall compression time.
For decompression, we employed Horner's method to evaluate the polynomial coefficients for reconstructing each vector. Here, OpenMP was again used with \texttt{\#pragma omp parallel for}, enabling each thread to independently compute the decompressed results for its assigned subset of vectors. By splitting the workload evenly across the 48 threads, we ensured that both phases of Ares—compression and decompression—fully utilized the available hardware resources, achieving high scalability and efficient runtime performance on modern multi-core systems.

For comparison, the following baseline algorithms were implemented and evaluated alongside Ares:

\begin{itemize}
    \item \textbf{PCA (Principal Component Analysis):} PCA was implemented using the Scikit-learn~\cite{scikit-learn} library, leveraging its optimized \texttt{fit\_transform} method for compression. The decompression step utilized the learned components to reconstruct the data via the \texttt{inverse\_transform} function. PCA inherently benefits from efficient matrix operations but does not natively support data parallelism or streaming.

    \item \textbf{NMF (Non-negative Matrix Factorization):} NMF was also implemented using Scikit-learn~\cite{scikit-learn}, employing the high-performance \texttt{fit\_transform} method with the multiplicative update solver. The decomposition into the \(W\) and \(H\) matrices allowed for reconstruction during decompression by computing their product. Like PCA, NMF relies on efficient matrix operations but can be computationally intensive due to its iterative nature.

    \item \textbf{Autoencoder:} The Autoencoder was implemented using PyTorch~\cite{pytorch}. A simple feed-forward neural network architecture was used, with the encoder reducing the input dimensionality to the target dimension and the decoder reconstructing the data. The model was trained using the mean squared error loss function. While training can be parallelized on GPUs, the inference phase for compression and decompression was executed on CPUs in our experiments.
\end{itemize}

Each algorithm was optimized to the extent possible under its respective implementation constraints, ensuring a fair comparison with Ares.

\subsection{Experimental Setup}

In this section, we describe the datasets used in our experiments and the platform on which the evaluations were conducted. These details ensure reproducibility and provide context for interpreting the results.

\subsubsection{Datasets}

We selected a diverse set of datasets to evaluate the performance of Ares and baseline algorithms across various scenarios:

\begin{itemize}
    \item \textbf{Random Dataset:} A randomly generated dataset consisting of 1,000 samples with 1,000 features each. This dataset serves as a baseline for understanding the behavior of algorithms on high-dimensional data without inherent structure.

    \item \textbf{Malicious URL Dataset~\cite{ma2009beyond}:} This dataset contains feature vectors derived from URL attributes for classifying benign and malicious URLs. It evaluates the ability of compression algorithms to handle feature-rich and structured datasets with real-world security applications.

    \item \textbf{Newsgroups Dataset~\cite{lang1995newsweeder}:} A dataset of high-dimensional sparse vectors representing text documents across 20 categories. This dataset tests the performance of algorithms on sparse, text-based data.

\end{itemize}

    
\subsubsection{Platform}

All of the experiments were conducted on the NSF Chameleon Cloud~\cite{keahey2020lessons} platform, utilizing a high-performance compute node. The machine is powered by two Intel Xeon E5-2670 v3 CPUs, operating at a base clock speed of 2.30 GHz. Together, the processors provide a total of 48 hardware threads. The CPU features a hierarchical cache structure, including 32 KB L1 data and instruction caches, 256 KB L2 caches per core, and a shared 30 MB L3 cache per processor. These specifications ensure efficient handling of computation-intensive tasks.

The system is equipped with 128 GiB of RAM, sufficient for processing large datasets and performing memory-intensive operations associated with compression and decompression experiments. Storage is provided by a 250 GB Seagate SATA SSD, which offers low-latency access to data files.

For GPU-based computations, the node includes two NVIDIA Tesla P100 GPUs, renowned for their high parallel processing capabilities. However, the current implementation of Ares focuses solely on CPU-based execution, leaving GPU acceleration as a promising direction for future optimization.

\subsubsection{Parameter Configuration}

This subsection details the key parameters and configurations used in our experiments to ensure clarity and reproducibility.

\paragraph{Data Dimensionality and Normalization.}  
All datasets were standardized to have an original dimensionality of 1,000 features. When datasets exceeded 1,000 samples, random subsampling was applied; datasets with fewer than 1,000 features were zero-padded. Input data for all algorithms was normalized to the range \([0, 1]\).

\paragraph{Compression Parameters.}  
The target dimensionality for all algorithms was set to 10 by default, representing a 100x compression ratio. For Ares, this corresponded to using polynomials of degree 9. Autoencoders used a single hidden layer with size matching the target dimensionality, trained using the Adam optimizer (learning rate: 0.001) over 50 epochs. PCA employed full singular value decomposition (SVD), and NMF used a non-negative least squares solver, with default hyperparameters from Scikit-learn.

\paragraph{Repetition and Timing.}  
Each experiment was repeated 5 times to account for variability, with results averaged across repetitions. Compression and decompression times were measured using Python's \texttt{time} module, ensuring precision to milliseconds. Algorithms supporting parallel execution (e.g., Ares and PCA) leveraged the full 48 hardware threads available on the experimental platform.

\subsection{Compression Time}

This experiment aims to evaluate and compare the computational efficiency of four algorithms—Ares, PCA, NMF, and Autoencoder—by measuring their compression times on three datasets: Random, URL, and Newsgroup. The Random dataset simulates unstructured high-dimensional data, while the URL and Newsgroup datasets represent more structured, real-world scenarios. For each algorithm, we compress the original 1000-dimensional vectors into a target dimension of 10. The compression time was recorded using a high-performance computing platform with 48 hardware threads to ensure fair comparison across all algorithms. The Y-axis of the figure is logarithmic to better illustrate the large variations in computational times across algorithms.

\begin{figure}[t!]
\centering
\includegraphics[width=\columnwidth]{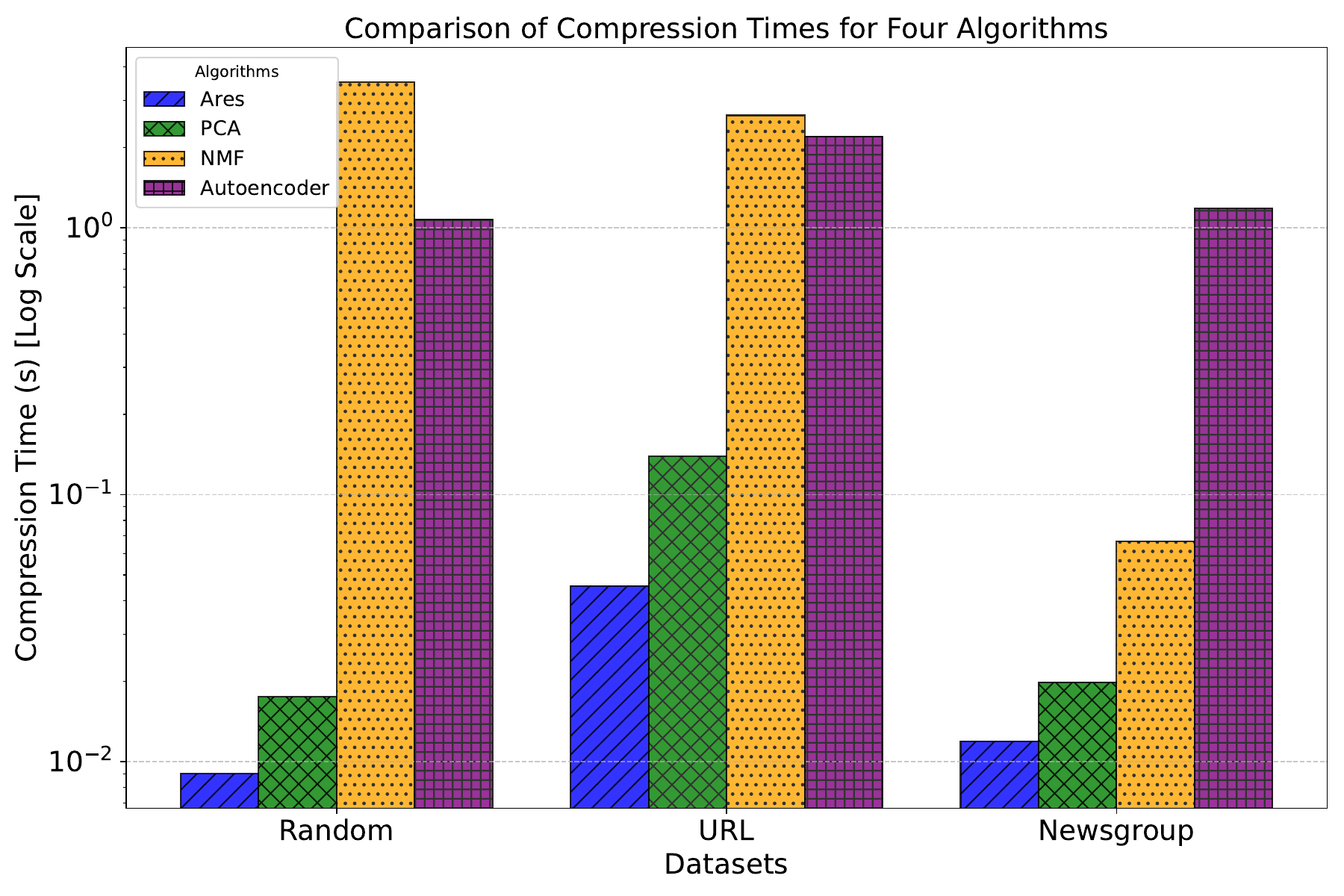}
\caption{Comparison of Compression Times for Four Algorithms on Random, URL, and Newsgroup Datasets (Target Dimension: 10). Note: The Y-axis is in logarithmic scale for better visualization of differences.}
\label{fig:compression-times}
\end{figure}

Figure~\ref{fig:compression-times} illustrates the compression times of four algorithms—Ares, PCA, NMF, and Autoencoder—across three datasets: Random, URL, and Newsgroup. The Y-axis represents the compression time in seconds on a logarithmic scale, providing a clear comparison of performance differences across varying magnitudes. The results show that Ares consistently achieves the lowest compression time across all datasets, demonstrating its efficiency compared to other methods. PCA exhibits moderate performance, outperforming NMF and Autoencoder but lagging behind Ares. NMF and Autoencoder, while effective in certain applications, incur significantly higher compression times, particularly for larger or more structured datasets like URL and Newsgroup. Notably, Autoencoder consistently has the highest compression time due to the computational overhead associated with neural network training.


\subsection{Compression Ratio}

In this experiment, we compare the compression ratios of four algorithms—Ares, PCA, NMF, and Autoencoder—across three datasets: Random, Malicious URL, and Newsgroup. The compression ratio is calculated as the ratio of the original data size to the compressed representation size, expressed as a percentage. This metric reflects the storage efficiency achieved by each algorithm. The datasets were selected to capture different characteristics: Random provides synthetic, unstructured data; Malicious URL focuses on real-world security-related features; and Newsgroup represents textual, sparse, and high-dimensional data.

\begin{figure}[t!]
    \centering
    \includegraphics[width=\columnwidth]{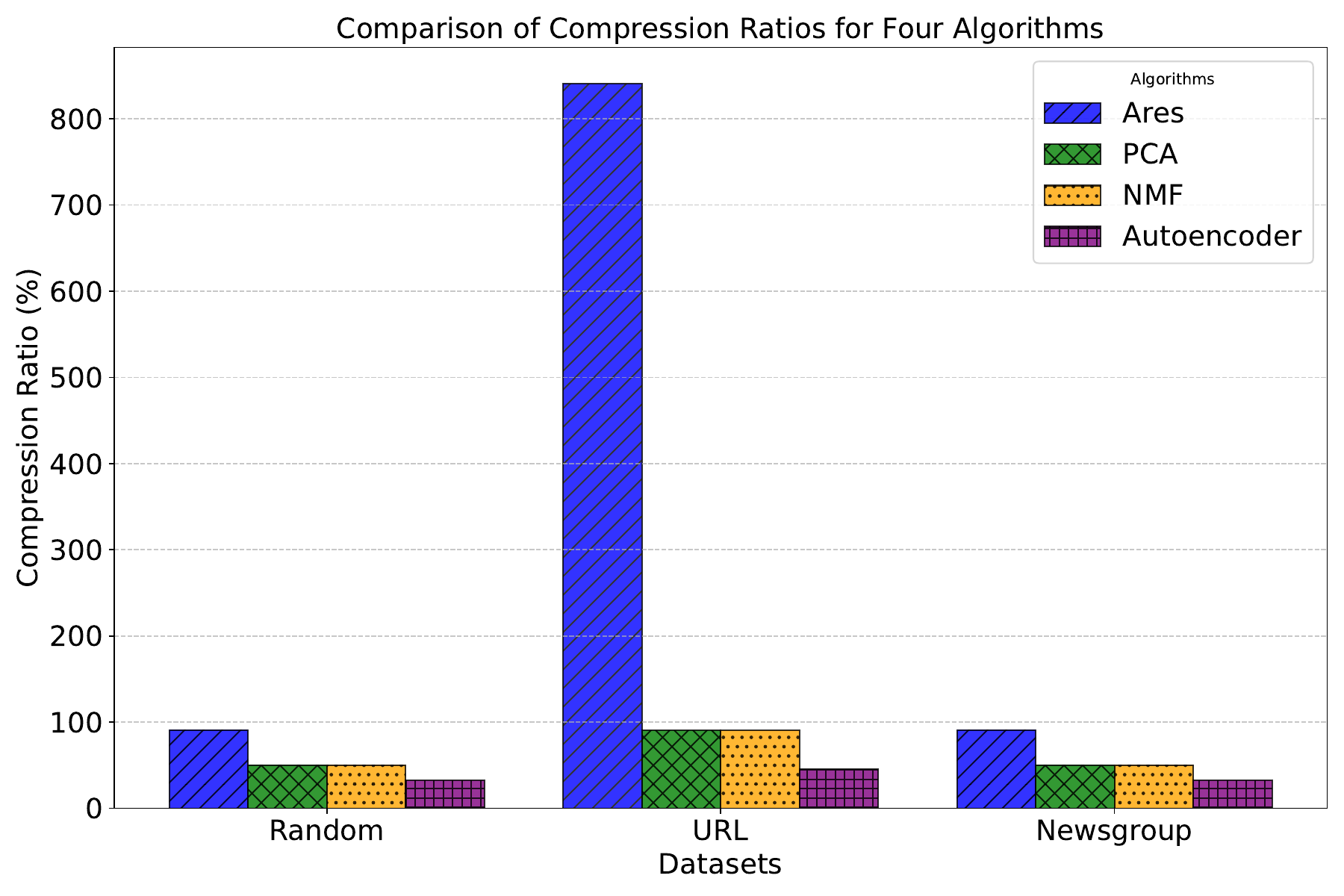}
    \caption{Comparison of Compression Ratios for Four Algorithms on Random, URL, and Newsgroup Datasets (Target Dimension: 10). Ares demonstrates superior compression efficiency on the sparse URL dataset but comparable performance on other datasets.}
    \label{fig:compression-ratio}
\end{figure}

The results, presented in Figure~\ref{fig:compression-ratio}, show that Ares achieves the highest compression ratio for the URL dataset, surpassing 840\%, due to the dataset's sparsity and Ares's ability to leverage polynomial representations. However, for the other two datasets, Ares's compression ratio is comparable to PCA and NMF. The Autoencoder, while effective in preserving reconstruction quality, shows lower compression ratios due to the need for additional parameters in its neural network architecture. These results highlight Ares's strength in handling sparse, structured data, but also indicate that its efficiency depends on the data distribution and underlying sparsity. Further optimizations may enhance its performance for denser datasets like Random and Newsgroup.


\subsection{Decompression Time}

This experiment evaluates the decompression times for four algorithms—Ares, PCA, NMF, and Autoencoder—on three datasets: Random, URL, and Newsgroup. The decompression phase is critical for real-world applications, as it determines the efficiency of restoring compressed data to its original form. The time taken to decompress data is measured in milliseconds (ms) for consistency, providing a fine-grained comparison across the algorithms and datasets.

\begin{figure}[t!]
\centering
\includegraphics[width=\columnwidth]{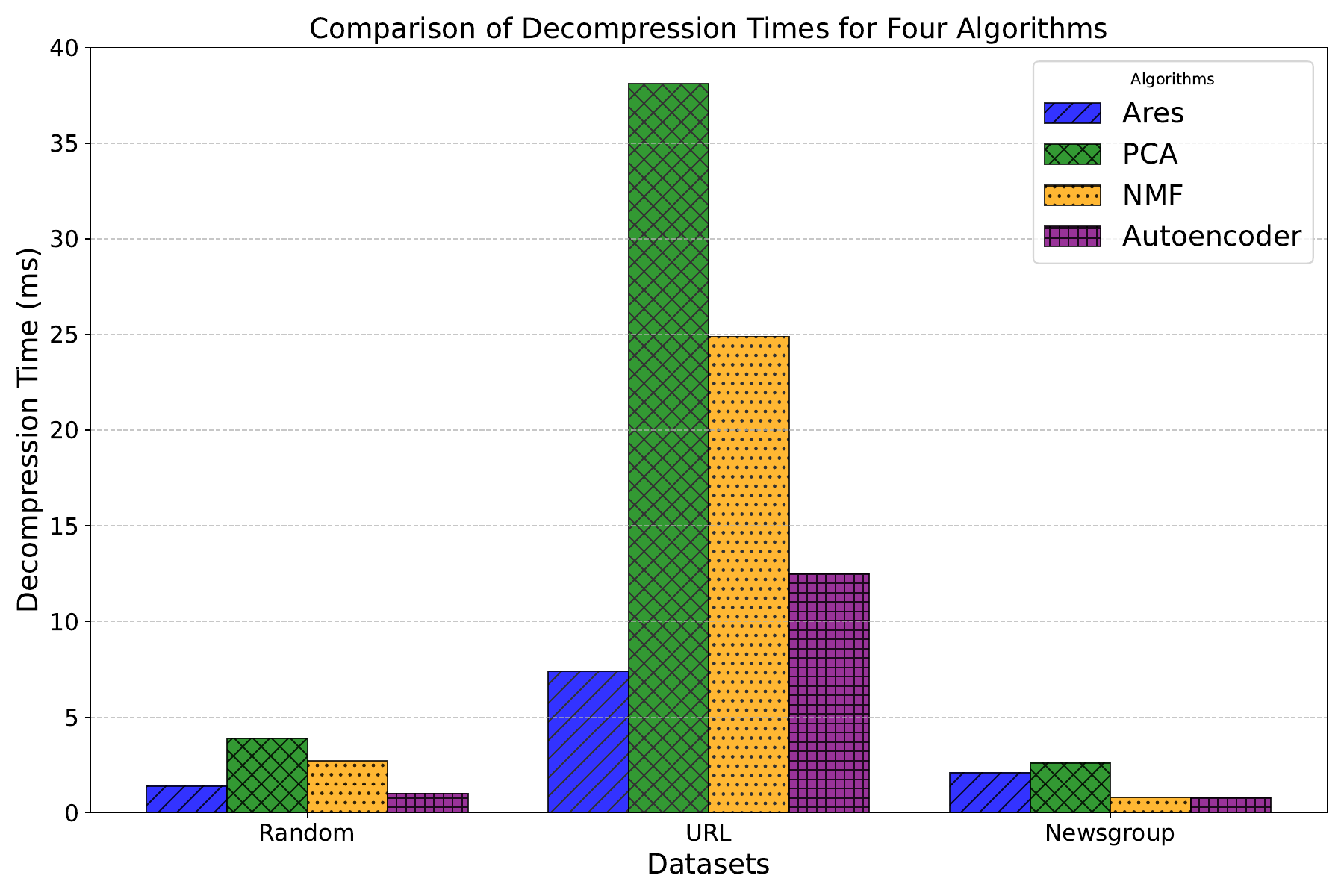}
\caption{Comparison of Decompression Times for Four Algorithms on Random, URL, and Newsgroup Datasets (Target Dimension: 10). The time is measured in milliseconds (ms).}
\label{fig:decompression-time}
\end{figure}

As shown in Figure~\ref{fig:decompression-time}, the Ares algorithm demonstrates competitive decompression times, particularly on the Random and URL datasets, where it outperforms PCA, NMF, and Autoencoder. However, on the Newsgroup dataset, Ares has slightly longer decompression times compared to NMF and Autoencoder.
The reason for this is that the Newsgroup dataset has a simpler and more regular feature structure compared to the URL dataset. For simpler datasets like Newsgroup, the more complex factorization and reconstruction computations required by Ares result in marginally longer decompression times than the relatively simpler NMF and Autoencoder approaches.

It is important to note that the differences in decompression times are quite small, on the order of a few milliseconds. This suggests that the performance of all these algorithms is quite close on the Newsgroup dataset, and the choice between them should be based on factors beyond just decompression speed, such as compression ratio, memory usage, and the specific needs of the application.


\subsection{Reconstruction Quality}

Table~\ref{table:mae-comparison} presents a comparison of the Mean Absolute Error (MAE) across four different algorithms: Ares, PCA, NMF, and Autoencoder.
The Mean Absolute Error (MAE) is a widely used metric for evaluating the performance of predictive models. It is defined as the average of the absolute differences between the predicted values and the true (or observed) values. The direct interpretation of MAE is the average magnitude of the errors in the same unit as the original data. For example, if the target variable is in dollars, the MAE will also be in dollars, representing the average absolute difference between the predicted and true values. The intuitive meaning of a low MAE is that the predictions made by the model are, on average, close to the true values. Conversely, a high MAE indicates that the predictions deviate significantly from the true values on average.

\begin{table}[h!]
\centering
\caption{Comparison of MAE (Mean Absolute Error) Across Four Algorithms}
\label{table:mae-comparison}
\begin{tabular}{lccc}
\toprule
\textbf{Algorithm}    & \textbf{Random} & \textbf{URL} & \textbf{Newsgroup} \\
\midrule
Ares               & 0.25            & 0.02                  & 0.00               \\
PCA                & 0.24            & 0.01                  & 0.00               \\
NMF                & 0.24            & 0.01                  & 0.00               \\
Autoencoder        & 0.26            & 0.01                  & 0.00               \\
\bottomrule
\end{tabular}
\end{table}

The results indicate that on the Random dataset, the four algorithms exhibit relatively similar MAE values, with Ares achieving an MAE of 0.25, PCA and NMF both achieving 0.24, and Autoencoder performing slightly worse at 0.26. However, on the URL dataset, Ares, PCA, and NMF demonstrate slightly lower MAE values of 0.02, 0.01, and 0.01 respectively. Interestingly, on the Newsgroup dataset, all four algorithms achieve the same remarkable MAE of 0.00, indicating their ability to make highly accurate predictions on this particular dataset.
The sparsity of the Newsgroup data essentially makes the reconstruction task easier for these algorithms, as they can focus on accurately preserving the few non-zero elements rather than having to precisely model a dense, high-dimensional vector space. This property of sparse data contributes to the exceptional performance of the tested algorithms on this dataset.

\section{Conclusion}

This paper introduced Ares, a stateless compression framework for efficient processing of high-dimensional data. By leveraging polynomial representations, Ares achieves compact compression while supporting algebraic operations directly in the compressed domain. The stateless design eliminates auxiliary data, making it ideal for streaming and infinite datasets. Experiments on synthetic and real-world datasets show that Ares outperforms methods like PCA and Autoencoders in compression time and scalability, with comparable reconstruction accuracy. These results showcase the practicality of Ares for real-world applications involving large-scale, high-dimensional data.


\section*{Acknowledgments}

Results presented in this paper were partly obtained using the Chameleon testbed supported by the National Science Foundation. 

\bibliographystyle{unsrt}
\bibliography{ref_new}

\end{document}